\documentclass[a4paper,conference]{IEEEtran}
\IEEEoverridecommandlockouts
\usepackage{times}
\usepackage{epsfig}
\usepackage{graphicx}
\usepackage{amsmath}
\usepackage{amssymb}
\usepackage{multirow}
\usepackage{float}
\usepackage{placeins}
\usepackage[sort]{cite}
\usepackage{amsmath}
\usepackage{etoolbox}
\usepackage{bm}
\usepackage{amsthm}
\usepackage{mathtools}
\usepackage{algorithm}
\usepackage{algpseudocode}
\usepackage{xspace}

\newcommand{\MyMapTemplateNoPrefix}[3]{\expandafter#1\csname#3\endcsname{#2{#3}}}
\forcsvlist{\MyMapTemplateNoPrefix {\def} {\mathbf} } {0,1,a,b,c,d,e, f, g, h, i, j, k, l, m, n, o, p, q, r, u, v, w, x, y, z} % e.g., \a --> \mathbf{a}
\newcommand{\MyMapTemplatePrefix}[4]{\expandafter#1\csname#3#4\endcsname{#2{#4}}} % it remembles a template: \#3#4 --> #2{#4}
\forcsvlist{\MyMapTemplatePrefix {\def} {\mathcal}{c}} {A,B,C,D,E,F,G,H,I,J,K,L,M,N,O,P,Q,R,S,T,U,V,W,X,Y,Z}  % e.g., \cA --> \mathcal{A}
\forcsvlist{\MyMapTemplatePrefix {\def} {\mathbf} {b}} {A,B,C,D,E,F,G,H,I,J,K,L,M,N,O,P,Q,R,S,T,U,V,W,X,Y,Z} % e.g, \bA --> \mathbf{A}
\forcsvlist{\MyMapTemplatePrefix {\DeclareMathOperator} {} {} } {trace, argmax, argmin, MixUp, Shuffle, Concat, aggregate, features, Freeze, Unfreeze, mAP, Melt}

\def\etal{\emph{et al.}\@\xspace}

\ifCLASSINFOpdf
\else
\fi

\begin{document}

\title{When Few-Shot Learning Meets Video Object Detection}

% \author{\IEEEauthorblockN{Michael Shell}
% \IEEEauthorblockA{School of Electrical and\\Computer Engineering\\
% Georgia Institute of Technology\\
% Atlanta, Georgia 30332--0250\\
% Email: http://www.michaelshell.org/contact.html}
% \and
% \IEEEauthorblockN{Homer Simpson}
% \IEEEauthorblockA{Twentieth Century Fox\\
% Email: homer@thesimpsons.com}
% \and
% \IEEEauthorblockN{James Kirk\\ and Montgomery Scott}
% \IEEEauthorblockA{Starfleet Academy\\
% San Francisco, California 96678--2391\\
% Telephone: (800) 555--1212\\


\author{\IEEEauthorblockN{Zhongjie Yu\IEEEauthorrefmark{1}\IEEEauthorrefmark{2},
Gaoang Wang\IEEEauthorrefmark{3},
Lin Chen\IEEEauthorrefmark{1}\IEEEauthorrefmark{5},
Sebastian Raschka\IEEEauthorrefmark{2} and
Jiebo Luo\IEEEauthorrefmark{4}}
\IEEEauthorblockA{\IEEEauthorrefmark{1}Wyze Labs, Inc.}
\IEEEauthorblockA{\IEEEauthorrefmark{2}University of Wisconsin-Madison, USA}
\IEEEauthorblockA{\IEEEauthorrefmark{3}Zhejiang University, China}
\IEEEauthorblockA{\IEEEauthorrefmark{4}University of Rochester, USA}
\IEEEauthorblockA{\IEEEauthorrefmark{5}Email: gggchenlin@gmail.com}}
% \thanks{\IEEEauthorrefmark{5}Corresponding author: Lin Chen. Email: gggchenlin@gmail.com}


\maketitle


\begin{abstract}
Different from static images, videos contain additional temporal and spatial information for better object detection. However, it is costly to obtain a large number of videos with bounding box annotations that are required for supervised deep learning. Although humans can easily learn to recognize new objects by watching only a few video clips, deep learning usually suffers from overfitting. This leads to an important question: how to effectively learn a video object detector from only a few labeled video clips?
In this paper, we study the new problem of few-shot learning for video object detection. We first define the few-shot setting and create a new benchmark dataset for few-shot video object detection derived from the widely used ImageNet VID dataset. We employ a transfer-learning framework to effectively train the video object detector on a large number of base-class objects and a few video clips of novel-class objects. By analyzing the results of two methods under this framework (Joint and Freeze) on our designed weak and strong base datasets, we reveal insufficiency and overfitting problems. A simple but effective method, called Thaw, is naturally developed to trade off the two problems and validate our analysis. 
 Extensive experiments on our proposed benchmark datasets with different scenarios demonstrate the effectiveness of our novel analysis in this new few-shot video object detection problem.
\end{abstract}


% \fi
\IEEEpeerreviewmaketitle

\FloatBarrier

\section{Introduction}
\label{sec:intro}

% 想表达的意思是

% 提出问题： 如果从few clips 去学习到一个好的video object detector. 又要涉及到video特别性，又要讲到few-shot然后引出小样本学习概念。

% 小样本学习发展主要在image classification上，只有很少work on  object detection 但也都是image上的domain.

% 而video的work也比较少，主要是探究.并且对于video的小样本的定义也没有 （image 的定义很直接，1shot is 1 object; image classification 就是一张图，image object detection 就是一个物体和他的bounding box）。

% 基于这些问题和motivation. 我们率先定义并研究了few-shot video object detection. 提出使用few-shot learning 中transfer learning 的框架，并且分析了一些情境下使用不同方法的效 under our proposed framework果，基于发现 我们提出了insufficient 和overfitting problem，我们purpose 一种新的方法，simple but effective to balance insufficient and overfitting problem. extensive

% Although recent years have witnessed the tremendous success of applying deep convolutional neural networks (CNNs) for image object detection, . On the other hand, these methods easily suffer from the overfitting problem when there are only a few samples for novel-class objects.

% Learning new concepts from a few samples is known as few-shot learning. It is investigated a lot in the past few years 

% Humans could learn to re. Nevertheless, Deep

With the popularity of cameras in surveillance systems and mobile phones, as well as the mass adoption of social media content sharing platforms, there are more and more video content generated every day. Therefore, the need for developing algorithms to detect objects in videos grows rapidly in computer vision. Although it is possible to train a robust video object detector with sufficient labeled videos, powerful deep neural networks and abundant computational resources, collecting such a large number of videos with bounding box annotations is costly.

\FloatBarrier
\begin{figure*}[tbp]
    \centering
    \includegraphics[width=0.95\textwidth]{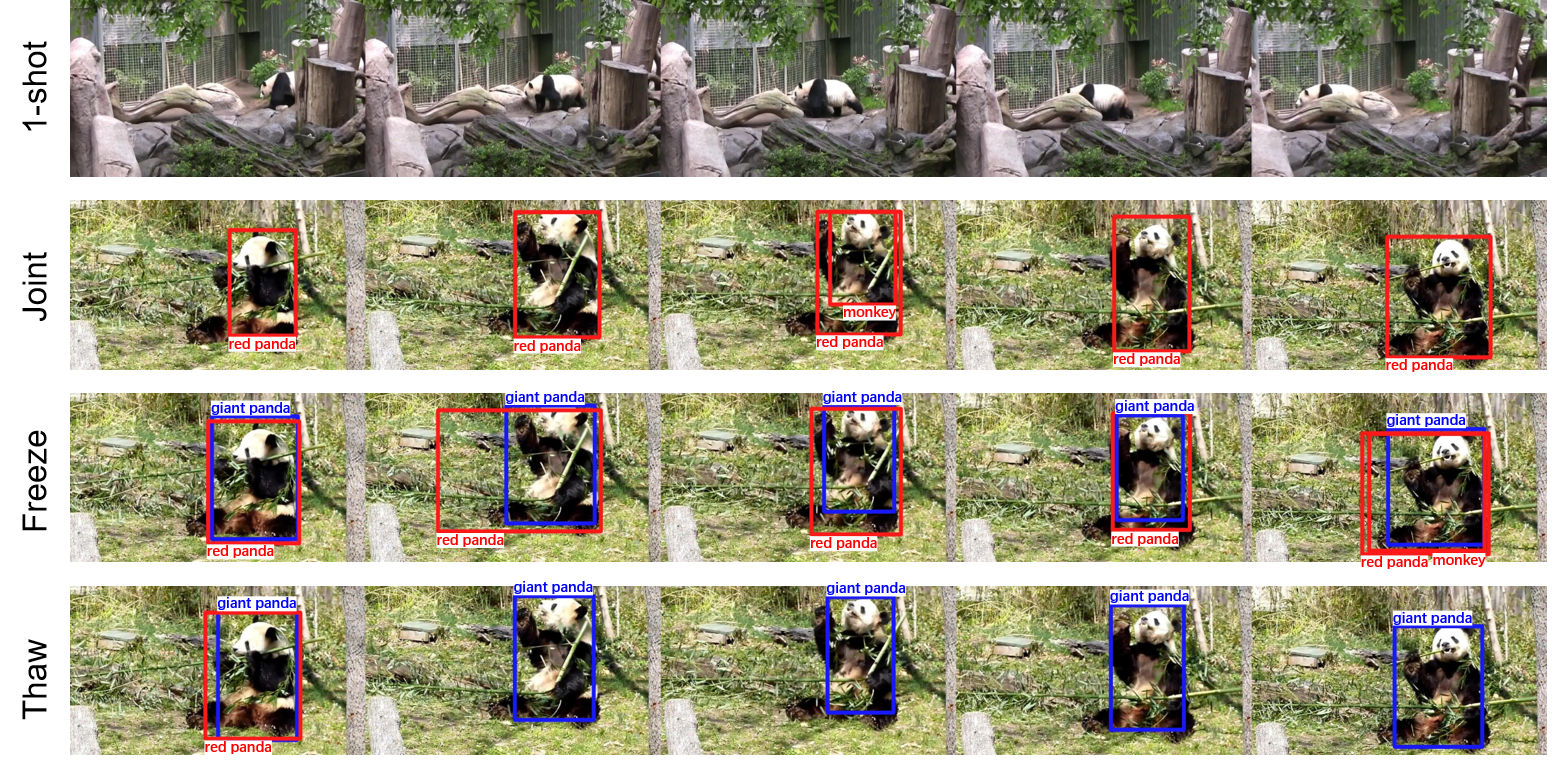}
    \caption{Examples of few-shot learning for video object detection of {\it giant panda} (weak base dataset). Blue (\emph{resp.} red) bounding boxes denote  correct (\emph{resp.} incorrect) detection. Note that {\it red panda} is a {\it different} animal.  The 1st row shows the 1-shot novel-class video used in few-shot adaptation. The 2nd, 3rd, and 4th rows show the detection results in the validation videos by Joint, Freeze and Thaw, respectively. }
    \label{fig:panda}
              \vspace{-0.2in}

\end{figure*}

Humans can learn new concepts easily with only a few examples. Despite that deep learning has been successfully applied to many real-world applications, it usually suffers from the overfitting problem when there are only a few samples for new concepts. Few-shot learning, which tries to learn a robust model from only a few samples of a new concept, has thus attracted great attention recently \cite{imprinting,matchingnetwork,maml,prototypical,LEO,TPN,DN4,reptile,feature-reweight,repmet,attention-rpn,incremental,simple,video-classification,video2,transmatch}. 

Most existing few-shot learning methods focus on either image classification \cite{imprinting,maml,activation,matchingnetwork,prototypical,LEO,TPN,DN4,reptile} or video classification \cite{video-classification,video2}. While some recent few-shot learning methods \cite{feature-reweight,repmet,attention-rpn,incremental,simple} have investigated object detection, all of them focus on object detection in static images instead of videos. Different from static images, videos contain abundant spatial and temporal information of objects. %It is said that ``a picture is worth a thousand words'', and even a short video contains lots of frames and thus much more information than a picture. 
Therefore, it becomes more  imperative to design a model for video object detection given a few videos of novel-class objects. This poses a new problem of {\it few-shot learning for video object detection}.

Since videos contain more information than static images, using the spatial-temporal information in videos is critical to achieving good performance.  However, since it is {\bf computationally prohibitive} to build episodes by representing a video by its frames, the {\bf meta-learning methods cannot be directly used to solve our few-shot video object detection problem}. While some techniques target few-shot video classification \cite{video-classification,video2}, it is different from video object detection, as video classification only aims to classify the entire video as one of the classes. In contrast, video object detection needs to detect both the presence and spatial-temporal location of an object in all frames in a video. This increases difficulty exponentially in terms of both computing and realization. The recent finding \cite{simple} that `frustratingly simple' transfer-learning based methods %(\ie, freezing feature extractor when fine-tuning )
can achieve good results on image object detection opens a possible path for solving the problem of few-shot video object detection. We are interested that {\bf if it is also `frustratingly simple' when few-shot learning meets video object detection}?

Our contributions in this paper are summarized as follows:

\noindent 1) We propose a new paradigm of few-shot learning for video object detection. Specifically, we study how an object detector can be learned from a few videos of new concepts where abundant temporal information is available while maintaining good performance for existing classes.
    
\noindent 2) Realizing there is no prior work studying this, we curate a dataset derived from the popular ImageNet-VID dataset \cite{ILSVRC15} for investigating this new problem. A strong base dataset and a weak base dataset are designed for further scenario analysis.
    
\noindent 3) We adopt a transfer-learning framework for solving this problem and investigate two methods under the framework: Joint and Freeze. We reveal two issues: {\it insufficiency} and {\it overfitting} based on a novel quantitative analysis.
    
\noindent 4) We propose a simple method called Thaw to trade off the {\it insufficiency} and {\it overfitting} problems revealed by our analysis. Extensive experiments demonstrate that our proposed Thaw naturally motivated by our novel analysis can help the video object detector efficiently learn new concepts from a few novel-class videos and achieve promising performance.

\section{Related Work}

\subsection{Few-Shot Learning}

Few-shot learning methods can be categorized into meta-learning based methods and transfer-learning methods.

\noindent{\bf Meta-learning} aims to learn a paradigm based on the base-class data with an episodic training strategy so that it can generalize to new tasks with only a few novel examples \cite{matchingnetwork,prototypical,relationnet,DN4,maml,LEO}. 
% Metric-based meta-learning methods learn a good distance metric from the few-shot examples of base classes \cite{matchingnetwork,prototypical,relationnet,DN4}. For example, Prototypical Network \cite{prototypical} measures the distance from queries and the prototypes of each class. %Relation Net \cite{relationnet} employed a network to learn the similarities between the support and query sets.
% DN4 \cite{DN4} explores the local descriptors to measure similarities. Optimization-based meta-learning methods like MAML \cite{maml}, LEO \cite{LEO} and  Reptile \cite{reptile} aim to find a good optimization direction that converges faster to the optimal solution with fewer gradient steps.
However, most meta-learning based methods are designed for image classification. It is challenging to directly extend them to the few-shot video object detection scenario in this paper.
\noindent{\bf Transfer-learning} based methods focus on how to train a good base model from the large amount of base-class data and then adapt the model to novel classes with only a few-shot of samples \cite{imprinting,activation,closerlook}. Unlike meta-learning based methods, the base model is trained using traditional methods and a new classifier is built with a {\bf frozen} feature extractor. This `simple' method could achieve comparable or even better performance compared with popular but complicated meta-learning methods revealed by Chen \etal \cite{closerlook}.

\subsection{Few-Shot Image Object Detection}

Few-shot image object detection is is still a developing area of research, with only a handful of notable works  \cite{meta-rcnn,feature-reweight,repmet,attention-rpn,incremental,simple}. Most of them are based on complicated meta-learning \cite{meta-rcnn,feature-reweight,attention-rpn}.
% Meta-RCNN \cite{meta-rcnn} proposed the meta-learning process for RoI features in Faster-RCNN \cite{faster-rcnn}. Feature Reweighting \cite{feature-reweight} developed a reweighting module and map features with corresponding classes in YOLOv2 \cite{yolov2}. Qi \etal \cite{attention-rpn} proposed Attention RPN and Multi-Relation Head for matching classes.
Recently, Wang \etal \cite{simple} investigated the ` frustratingly simple' transfer-learning based method by first training a Faster-RCNN \cite{faster-rcnn} model on the base-class data, and then freezing the feature extractor and fine-tuning a cosine classifier and a regressor on a balanced dataset of base-class and novel-class data. This method has achieved promising performance compared to previous meta-learning methods.

\subsection{Video Object Detection}

Object detection in videos is a challenging task due to the high variation across the videos. The objects in videos may be blurry  and change pose and status. Meanwhile, the moving background and unstable lighting conditions pose challenges to object detection. On the other hand, the temporal information in videos can provide more information than static images.
These aspects have remained under-explored in the past several years but have recently started to attract more attention after the ImageNet-VID dataset \cite{ILSVRC15} was released.
 Most recent works \cite{RDN,FGFA,STSN} focus on utilizing nearby frames in a short range to gather local information. Meanwhile, other works aim to use global information from frames in a wider range \cite{SELSA}. MEGA \cite{MEGA} was recently proposed to effectively combine the local and global information so that one frame can access more content from nearby and far-away frames, resulting in state-of-the-art performance.

\section{Few-Shot Video Object Detection}

In this section, we introduce the definition of {\it few-shot video object detection} and dataset construction process.

\FloatBarrier

\begin{figure*}[tp]
    \centering
    \includegraphics[width=0.8\textwidth]{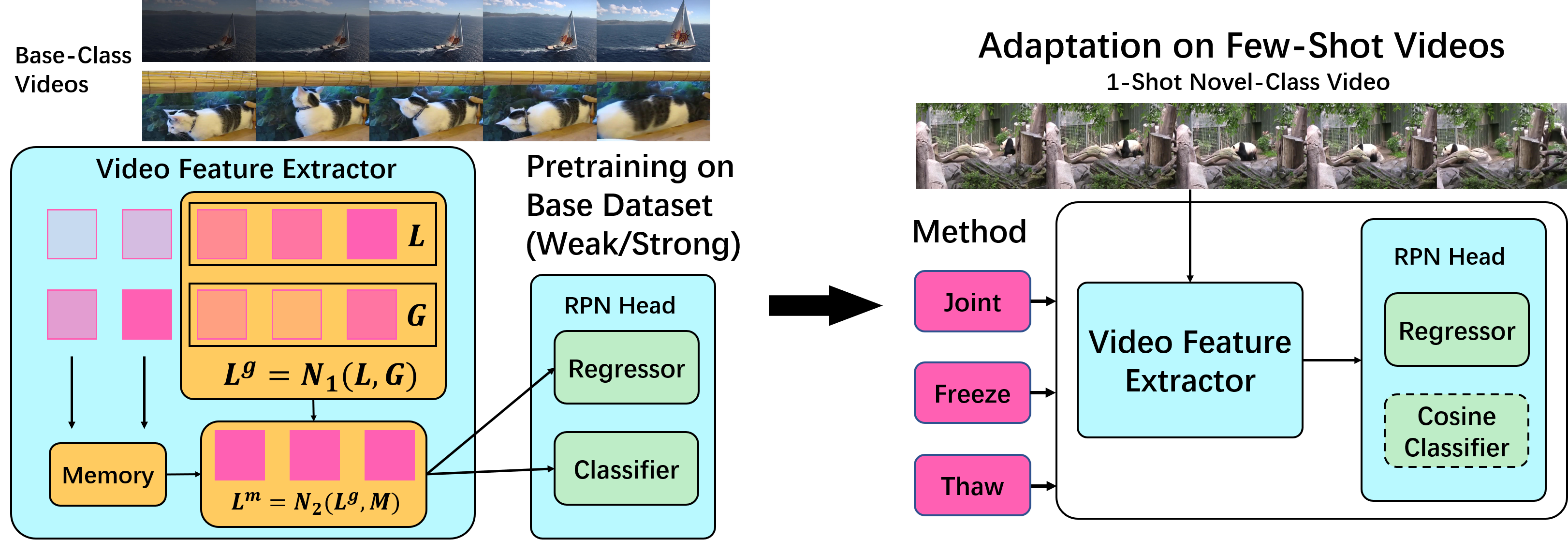}
    \caption{The proposed framework of few-shot learning for video object detection. A video object detector is pretrained on the base dataset by aggregating local and global information from different frames in the videos, and then adapted to novel classes based on few-shot novel-class video dataset. During the adaptation, the cosine classifier is used in the detection head and the model is fine-tuned by three methods: Joint, Freeze, and our developed Thaw.}
    \label{fig:process}
              \vspace{-0.2in}

\end{figure*}

\subsection{Problem Definition}

Let us define two sets of classes: base classes $\cC_{base}$ and novel classes $\cC_{novel}$, where $|\cC_{base}|=M$, $|\cC_{novel}|=N$ with $\cC_{base}\cap \cC_{novel}=\emptyset$. Assume there is a large base dataset consisting of 
many videos per class $\cV_{base}=\{V_{base}^i|i=1,\dots\}$, where $V_{base}^i=\{(I_1^i, Y_1^i),(I_2^i, Y_2^i),\dots\}$ is the $i$-th video in the base dataset. $I_t^i$ is the $t$-th frame in the $i$-th video, and $Y_t^i=\{(c_{t,1}^i,B_{t,1}^i),(c_{t,2}^i,B_{t,2}^i),\dots \}$ with $c_{t,b}^i$, $B_{t,b}^i$ being the class and the bounding box coordinates for $b$-th object in 
$t$-th frame of the $i$-th video, respectively. Note that $Y_t^i$ could be $\emptyset$ and $c_{t,b}^i \in \cC_{base}$.
A video object detection model can be built on $\cV_{base}$.

After building the model from the base dataset, we would like to adapt the base model to novel classes given only a few videos per novel class. It is natural to define {\em shot} for image classification and image object detection because one image or one bounding box has only one class label. In order to have an appropriate definition of {\em shot} for video object detection, we define two types of videos as follows.

\noindent{\bf Clean videos:} Videos contain only one class of objects. Each frame may contain more than one object of that class.

\noindent{\bf Perfect videos:} Videos contain only one class of objects and each frame contains only one object from that class. It is a subset of clean videos.

We denote one perfect video as one-shot. Also, we expect a good detector to perform well on both base and novel classes. Motivated by these aspects, we denote an $(N+M)$-way $K$-shot balanced dataset with both novel and base classes from perfect videos as $\cV_{balance}=\{V_{balance}^i | i=1,\dots,(N+M)\times K\}$, where $V_{balance}^i=\{(I_1^i, Y_1^i),(I_2^i, Y_2^i),\dots\}$, $Y_j^i={(c_{t,1}^i,B_{t,1}^i)}$ and $c_{t,1}^i \doteq c^i \in \cC_{novel}\cup \cC_{base}$ for all $t$. The model will be further trained on this balanced dataset.

%The model will be adjusted on this balanced $(N+M)$-way $K$-shot dataset.

% The meaning of selected these types of videos is to better define the $K$-shot situation in videos corresponding to the direct definition of $K$-shot in image classification and object detection.

% \section{The Proposed Framework for Few-Shot Video Object Detection}
% In this section, we introduce the framework for few-shot video object detection in 4-fold: 1) the problem definition, 2) dataset construction, 3) the modified state-of-the-art video object detector, 4) baseline methods for few-shot learning in video object detection.  
% Then, we are given a small few-shot ($N$-way $K$-shot) novel dataset $\cV_{novel}=\{V_{novel}^i,i=1,\dots\, N \times K \}$, where $V_{novel}^i=\{(I_1^i, Y_1^i),(I_2^i, Y_2^i),\dots\}$ is the $i$-th video in few-shot novel dataset. Here $Y_t^i={(c_{t,1}^i,B_{t,1}^i)}$ and $c_{t,1}^i \equiv c^i \in \cC_{novel}$ for all $t$, 

          \vspace{-0.05in}

\subsection{Dataset Construction}

% \begin{table}[]
%     \centering
%     \caption{The statistics of different types of datasets.} %\vspace{-0.05in}   
%     \begin{tabular}{c|c}
%     \hline
%          Dataset Type & Total Number of Frames  \\
%          \hline
%         All training videos      & 57,834 key frames\\
%     %   Clean training videos    & 43920 key frames \\
%       Perfect training videos &  31,530 key frames\\
%       All validation videos  & 176,126 frames\\
%       Balanced validation videos   & 23,624 frames\\
%       \hline
%     \end{tabular}

%     \label{tab:stat_video}
% \end{table}

Since we are studying a new few-shot learning video object detection problem, we construct a new dataset from ImageNet VID dataset \cite{ILSVRC15} widely used for video object detection. It consists of $30$ categories with $3862$ training videos and $555$ validation videos. We split them to 25 base classes and 5 novel classes in each novel-base split (A,B,C). Specifically, we create four different types of datasets as follows:

\noindent{\bf Strong base dataset:} It consists of all the videos from base-class objects. Meanwhile, like existing work for video object detection \cite{MEGA,leveraging}, we additionally include images from the image object detection dataset DET \cite{ILSVRC15}, leading to a {\it strong base dataset} for learning strong feature extractors of the base classes.

\noindent{\bf Weak base dataset:} It contains only the perfect videos from base classes, and thus is a subset of strong base dataset.

\noindent{\it Remarks:} We need to clarify that it is {\bf important} to investigate the impact from different types of base datasets for real-life applications since the available amount of data for base classes could vary in different scenarios. This has been overlooked in few-shot learning for image classification. Recently, Yue \etal \cite{interventional} investigated the different performances between strong and weak backbones used in few-shot image classification. This work shares similarities with our design for strong and weak base datasets.

\noindent{\bf Balanced few-shot dataset:} It is used for few-shot adaptation, and is randomly sampled from the perfect videos in the ImageNet VID training set for both novel and base class objects. We sample $K$ videos per class with $K=1,2,3$ as the shot number.

\noindent{\bf Balanced validation dataset:} It is the subset of the original clean validation videos and is used to evaluate the few-shot video object detector. Noting that the minimum number of clean validation videos among all classes is $3$, we randomly sample $3$ clean videos per class to construct this dataset. In this way, we can alleviate the impact from different numbers of videos in different classes.

% We consider to construct two types of base dataset: weak and strong. The video object detector would learn features on the two dataset thus extract weak and strong features.

% For a better description about how we build weak and strong dataset and the following few-shot dataset, we first define clean videos and perfect videos as following:

% \noindent{\bf Clean videos} are the videos where all frames belong to the same video that only contain one target class of objects. But each frame may contain more than one objects.

% \noindent{\bf Perfect videos} are subset of clean videos. All frames in one video contain one target class of objects and each frame only contains one object. 
% [may move to problem definition parts]

% Then the weak and strong base dataset is created as following:

% \noindent{\bf Weak base-class dataset} consists of perfect videos in training set for base-class objects. This would reduce the amount of data during training and build a weak feature extractor for us.

% \noindent{\bf Strong base-class dataset} consists of all videos containing base-class objects in training set. Meanwhile, as many work on video object detection \cite{MEGA,leveraging}, The images contain objects from same base classes in static image object detection dataset DET \cite{ILSVRC15} will also be included here to augment to learn better features.

% \noindent{\bf Balanced few-shot dataset} is constructed for fine-tuning and Thaw. We randomly sample from perfect videos in training set both on novel and base-class objects, with $K=1,2,3$ videos for each class.

\section{The Proposed Framework}

In this section, we introduce our proposed framework for few-shot video object detection, which is illustrated in Figure \ref{fig:process}. %Meta-learning is not suitable for our video object detection, as it requires a huge amount of memory to hold the episodes of support and query videos for object detection. Therefore, the convenient transfer-learning is employed in our framework.
First, a video object detector is pretrained on base dataset. Second, a cosine classifier is used to replace the classifier in the detection head and fine-tuned on the balanced few-shot video dataset. More details are provided in the subsequent sections.

\paragraph{Why not using meta-learning framework?}
Meta-learning methods are popular for image-level object detection, but it is computationally prohibitive or even unlikely to build episodes for thousands of frames from videos from different classes. This is special when few-shot learning meets video object detection.

\subsection{Part I: Pretraining the Video Object Detector}

In the first stage, we train a video object detector on the base dataset where many videos per class are provided. In particular, given the base dataset $\cV_{base}$, we first construct a video object detector to fully utilize the abundant video information by employing the state-of-the-art method MEGA \cite{MEGA}. Any video detector could be used here, and we use MEGA as it can efficiently combine both local and global information throughout the video.For each key frame $I_k$ in a given video, MEGA generates local pool $\cL$, global pool $\cG$, global aggregated pool $\cL^g$, enhanced pool $\cL^m$, memory module $\cM$ for ROI-Aligh and fully-connected layer to get the features.\footnote{For simplicity, we did not elaborate this in this paper.} The final features for all proposals is denoted as 
\begin{equation}
    f^e(I_k)=\left\{f^e(I_k)_1,f^e(I_k)_2,...\right\}.
\end{equation}
And $f^e(I_k)$ is used in the RPN head for classification and bounding box regression. This process is used for all the key frames in a video.

\subsection{Part II: Adaptation on Few-Shot Videos}

After obtaining the pretrained video object detector, we adapt the model based on the balanced few-shot video dataset including novel-class and base-class objects.

\subsubsection{Modification on the RPN Head} 
Since the cosine classifier is shown to be effective for few-shot image classification problem in \cite{closerlook}, and more suitable to de-correlate the feature space for different classes \cite{wang2017normface}, we adopt a cosine classifier for the few-shot fine-tuning stage. Specifically, a weight matrix  $\bW = [\w_1, ..., \w_{N+M}] \in  \cR^{d\times (N+M)}$ 
is fine-tuned where $\w_i$ represents the prototype of $i$-th class in $\cC_{novel} \cup \cC_{base}$. Then, the cosine similarity with the different classes for a given proposal $\x \coloneqq f^e(I_t)_l$ is written as:
\begin{eqnarray}
     S(\bW,\x) = \left[\cos(\theta(\w_1, \x), ..., \cos(\theta(\w_{N+M}, \x))\right]^\prime.
\end{eqnarray}
% Because cosine similarity ranges between -1 and 1, the softmax function is unable to predict the correct class via the one-hot encoding regime for class labels, causing a discrepancy between the one-hot and real distribution. In order to solve this issue, a scaling factor is usually applied on softmax for better convergence as used in \cite{imprinting,closerlook,wang2017normface}. With that, 
The probability for $i$-th class can be represented as:
\begin{equation}
    \frac{\exp(\sigma S(\bW,\x)_{i})}{\sum_c \exp(\sigma S(\bW,\x)_{c})},
\end{equation}
where $\sigma$ is the scale factor to solve the discrepancy between the one-hot and real distribution \cite{imprinting,closerlook,wang2017normface}.
 
% The structure of the regressor remains the same as in the pretraining stage but is appended $4\times N$ dimension (\ie,~coordinates of bounding boxes) to account for the $N$-way novel-class videos. The  weights of cosine classifier and regressor for novel-class videos are randomly initialized.

\subsubsection{Adaptation Strategies}

With the new RPN head,  we use multiple strategies (including the newly proposed {\bf Thaw} to be described in Section 6) to adapt the pretrained model when given few-shot videos.% including novel classes and base classes. 

\noindent{\bf Joint:} All the weights from the feature extractor and detection head are fine-tuned jointly on the {\it balanced few-shot dataset}. This joint fine-tuning is not designed for few-shot learning and usually suffers from overiftting, because the feature extractor can be easily impacted by the few-shot samples. Therefore, it is seldom used in few-shot image classification and usually serves as a {\bf low-performing baseline} for few-shot image object detection \cite{meta-rcnn,feature-reweight}.

\noindent{\bf Freeze:} In this method, the feature extractor is frozen and only the detection head is fine-tuned. Freezing feature extractor method is particularly suitable for few-shot learning and widely used in image classification \cite{closerlook}. The recent work in \cite{simple} also shows its simplicity and superiority in overcoming overfitting and achieves new state-of-the-art performance in few-shot image object detection.
%The purpose of doing that is to prevent the feature extractor impacted by the novel-class data and the classifier or detection head could be re-built by utilizing learnt features from novel-class data. 

\section{Experiments for Joint and Freeze}
In this section, we conduct experiments on the designed dataset under our proposed framework. We evaluate the different adaptation strategies of {\bf Joint} and {\bf Freeze} in different settings. We find it is not `frustratingly simple' when few-shot learning meets video object detection. We analyze the results and discover the {\it insufficiency} and {\it overfitting} problems in few-shot video object detection.

\subsection{Implementation Details}
\noindent{\bf Video object detector network:} ResNet-101 \cite{resnet} is used as the backbone. RPN head is applied to {\em conv4} block in ResNet where the anchors have $4$ scales and $3$ aspect ratios. $300$ bounding box proposals per frame are created during training and validation with the default IoU threshold set to $0.7$. Next,  RoI-Align and a fully-connected layer are employed  after {\em conv5} block to extract RoI pooled features, followed by the classifier and regressor. 
% For the video detector, we set the local pool and global pool range $T_l$ and $T_g$ to $12$ and 10, respectively. The hyperparameters in $\cN_1$ and $\cN_2$ modules are the same as in \cite{MEGA}.
For the cosine classifier, we set the scale factor $\sigma$ to $15$.

\noindent{\bf Training:} All models are trained on $4$ Tesla V100 GPUs, with each GPU holding one set of frames. 
During pretraining, the initial learning rate is set to $0.001$ and drops to $0.0001$ after $80,000$ iterations. We train the model for a total of $120,000$ iterations. During fine-tuning on the balanced few-shot dataset, the classifier and regressor are fine-tuned for $4,000$ iterations in all settings.  We set the learning rate during fine-tuning to $0.001$. We create three random splits (denoted as A, B, C) from novel-base classes, which is a common practice in few-shot image object detection. 
During training the detector on the base dataset and fine-tuning it on the balanced few-shot dataset, $15$ frames are evenly-spaced selected from the whole videos.
% For videos with fewer than $15$ frames, all frames are selected. 
The experiments for each few-shot video setting are repeated $5$ times. Each time the balanced few-shot dataset is selected {\bf randomly} but kept the {\bf same} for different methods to ensure a fair comparison. 
%As it is computational expensive on detecting videos and the shot in videos by our definition contains 15 frames of the same object, from our trying, using 5 times repeating could balance the time complexity and stability.
%  The code for the algorithm and dataset will be released.

\noindent{\bf Inference:} During inference on the validation set, we set the NMS threshold to $0.5$ IoU and use  mean average precision at 0.5 IoU (mAP50) as the evaluation metric. The balanced validation dataset remains the same for different few-shot settings to reduce randomness. The reported mAP50 in each setting is the average of $5$ random experiments.

\begin{table*}[tbp]
    \centering
      %  \fontsize{8}{9}\selectfont
    \caption{
    Novel-class mAP50 (in \%) on the validation videos when the base dataset is {\bf weak} or {\bf strong}. Better results are in bold. For novel-class performance, Freeze is {\bf better} than Joint on {\bf strong} base dataset but {\bf worse} than Joint on {\bf weak} base dataset, {\it opposite} to the research findings on images. }    \begin{tabular}{c|c|c|ccc|ccc|ccc}
    \hline
     Base  & \multirow{2}*{Class} & \multirow{2}*{Method / Split}  & \multicolumn{3}{c}{1-shot} & \multicolumn{3}{|c}{2-shot} & \multicolumn{3}{|c}{3-shot} \\
       %\cline{3-11}
    Dataset   & ~ & ~ & A & B & C & A & B & C & A & B & C \\
        \hline
      \multirow{2}*{\bf Weak}  & \multirow{2}*{Novel} & Joint (better) & {\bf 22.11} & 21.88 & {\bf 9.76} & {\bf 32.57} & {\bf 32.29} & {\bf 23.28} & {\bf 36.5} &  39.84 & {\bf 29.86}    \\
       ~ & ~ & Freeze & 18.93 & {\bf 24.05} & 8.09 & 25.07 & 31.60 & 13.88 & 28.89 & {\bf 40.53} & 13.73\\
        % \cline{2-12}
    %  ~    & \multirow{2}*{Base} & Joint & 56.41 & 59.23 & 56.30 & 58.41 & 61.77 & 58.86 & 60.12 & 63.75 & 60.26    \\
    %  ~ &  ~ & Freeze (better) & {\bf 60.98} & {\bf 63.97} & {\bf 60.95} & {\bf 61.15} & {\bf 64.12} & {\bf 60.83} & {\bf 61.32} & {\bf 65.02} & {\bf 61.10}  \\
        \hline
       \hline
        
       \multirow{2}*{\bf Strong} & \multirow{2}*{Novel} & Joint  &  21.69 & 31.10 & 20.06 & 39.37 & 45.94 & 34.74 & 44.56 & 51.43 & {\bf 43.33}    \\
        ~ & ~ & Freeze (better) & {\bf 41.85} & {\bf 40.69} & {\bf 31.71} & {\bf 50.14} & {\bf 47.81} & {\bf 42.66} & {\bf 53.15} & {\bf 52.41} &  43.08 \\
    %     \cline{2-12}
    %   ~ &  \multirow{2}*{Base} & Joint &   72.08 & 76.09 & 75.80 & 73.87 & 77.06 & 76.97 & 75.96 & 78.66 & 77.25 \\
    %   ~ & ~ & Freeze (better) & {\bf  82.79} & {\bf 84.75} & {\bf 86.45} & {\bf 83.00} & {\bf 84.69} & {\bf 86.53} & {\bf 83.14} & {\bf 85.00} & {\bf 86.43} \\
        \hline
    \end{tabular}

    \label{tab:weakandstrong}
              \vspace{-0.2in}

\end{table*}

\begin{figure}
    \centering
    \includegraphics[width=0.35\textwidth]{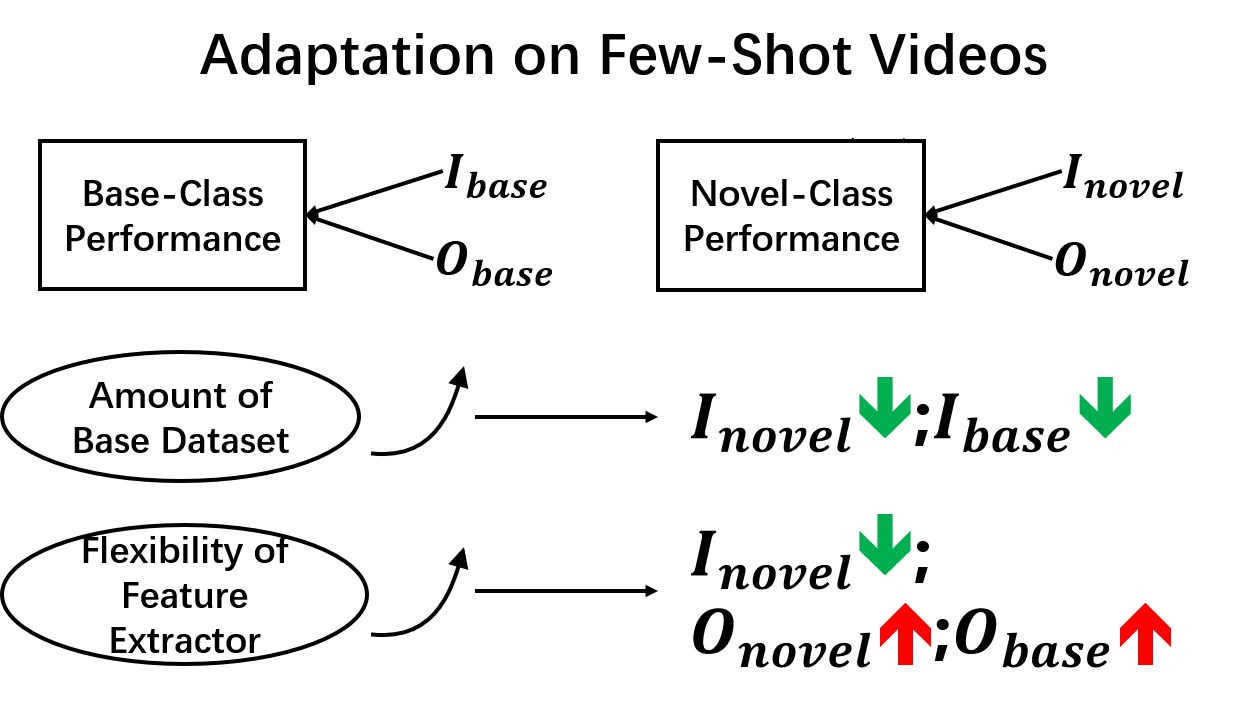}
    \caption{Illustration of the insufficiency and overfitting problems. $I$ represents the insufficiency problem and $O$ represents the overfitting problem. Strong and weak base datasets indicate the amount of base dataset. Freeze and Joint indicate the flexibility of the  feature extractor. The green down arrows indicate the problem is alleviated and the red up arrows indicate aggravated.}\vspace{-0.05in}
    \label{explain}
              \vspace{-0.15in}

\end{figure}

\subsection{Joint v.s. Freeze Results}

The experiment results for weak and strong base datasets are shown in Table~\ref{tab:weakandstrong}.

\noindent{\bf Novel-class performance:} When the base dataset is strong, {Freeze} is better than {Joint} on novel-class objects in almost all cases but one. This is consistent with the prior work in few-shot image classification and image object detection, as freezing the feature extractor can prevent overfitting. However, the pattern shown on the weak base dataset is opposite to strong base dataset (see Table \ref{tab:weakandstrong}). When the base dataset is weak, { Joint} generally performs better than { Freeze}, which is the {\it opposite} to the  the research findings on images. An explanation is that it is usually easier to obtain  sufficient information from the base dataset to build a sufficient feature extractor for the single image situation. Therefore, the overfitting problem will dominate for novel-class images and Freeze could work well in this situation.
However, the complicated structure and abundant information in videos may not be sufficiently learned from the base dataset and thus learning good features for novel-class objects cannot be guaranteed.

\subsection{Insufficiency vs. Overfitting}

The results from the previous section reveal the insufficiency and overfitting problems for the novel and base classes.

\noindent{\bf Insufficiency problem} corresponds to the situation that the features learned from the base dataset may not be sufficient for building detectors on novel-class objects.

\noindent{\bf Overfitting problem} corresponds to the situation that some good features learned from the base dataset may be distorted by the few-shot videos during fine-tuning when the feature extractor is unfrozen.
% a feature extractor learned from base class dataset performs very poorly when jointly fine-tuned with few-shot novel-class dataset when the feature extractor is unfrozen.

We summarize the analysis in Figure \ref{explain}. It is clear that a strong base dataset can alleviate the base-class insufficiency problem. Meanwhile, a strong base dataset can provide a better feature extractor, which improves the capability to extract better features for novel classes. Therefore, it can alleviate the insufficiency problem for novel classes.  On the other hand, freezing the feature extractor (Freeze) could largely solve the overfitting problem both for base and novel classes when fine-tuning on the few-shot videos. However, it does not allow the feature extractor to encode possible novel information from novel classes. Therefore,  unfreezing the feature extractor (Joint) would alleviate the insufficiency problem for novel classes since the feature extracted for base classes in the base training stage may not be sufficient to describe novel-class objects. 
% Since the base-class objects in few-shot videos do not provide any further useful information for the base classes (they come from the base-class objects used for pretraining the feature extractor), unfreezing the feature extractor (Joint) does not help reduce the base-class insufficiency problem. 

Therefore, in terms of novel-class performance, when the base dataset is \textbf{weak}, there is a significant novel-class insufficiency problem. Although Joint aggravates the novel-class overfitting problem, it largely alleviates novel-class insufficiency problem, such that its performance exceeds Freeze. When the base dataset is \textbf{strong}, the novel-class insufficiency problem becomes negligible, such that Freeze's performance is better than Joint's in this case. 

% On the other hand, in terms of base-class performance, unfreezing the feature extractor could only increase the base-class overfitting problem and could not reduce the base-class insufficiency problem, such that Joint's performance is always worse than Freeze no matter the base dataset is strong or weak.

\section{Experiments for Improved Method}

In this section, we further propose a simple but effective method called Thaw to balance the tradeoff of Joint and Freeze and demonstrate the rationality of our analysis.

\begin{table*}[tbp]
    \centering
    \caption{
    Novel-class and base-class mAP50 (in \%) on the validation videos, averaged from all novel-base splits. Best results are in bold. $^*$ indicates results are similar.}
    \begin{tabular}{c|c|ccc|c|ccc|c}
    \hline
       \multirow{2}*{Class} & \multirow{2}*{Method / Shot}  & \multicolumn{4}{c}{Weak Base Dataset} & \multicolumn{4}{|c}{Strong Base Dataset} \\
       %\cline{3-11}
        ~ & ~ & 1-shot & 2-shot & 3-shot & Rank & 1-shot & 2-shot & 3-shot & Rank \\
        \hline
        \multirow{3}*{Novel} & Joint  &  17.92 & 29.38 & 35.40 &2 & 24.28 & 40.02 & 46.44 &  3 \\
        ~ & Freeze & 17.02  & 23.52 & 27.72 & 3 & {\bf 38.08} & 46.87 & 49.55 & 1 $^*$ \\
        ~ & Thaw (Ours) & {\bf20.05} & {\bf 32.13} & {\bf 37.32} & 1 & 36.73  & {\bf 48.71} & {\bf 51.38} & 1 $^*$ \\
        \hline
         \multirow{3}*{Base} & Joint & 57.31 & 59.68 & 61.37 & 3 & 74.66 & 75.97 & 77.29  & 3 \\
        ~ & Freeze & {\bf 61.97} & {\bf 62.03} & {\bf 62.48} & 1 & {\bf 84.66} & {\bf 84.74} & {\bf 84.86}   & 1\\
        ~ & Thaw (Ours) & 60.13 & 60.33 & 61.52 & 2 &  80.79 & 78.81 & 78.94 & 2 \\ 
        \hline
    \end{tabular}
    
    \label{tab:Thaw}
              \vspace{-0.2in}

\end{table*}

          \vspace{-0.05in}

\subsection{Improved Method: Thaw}

As discussed in the previous section, there is a tradeoff between insufficiency and overfitting problems caused by unfreezing the feature extractor. To this end, we still first freeze the feature extractor and fine-tune on the detection head. This would give us a good detection head and prevent the overfitting problem. After convergence, we further unfreeze the feature extractor and fine-tune all the weights jointly to let the feature extractor learn extra information to reduce novel-class insufficiency problem. The detection head can be regarded as being initialized with better weights compared with Joint. We call this improved process {\em Thaw}:
\begin{equation}
\Freeze f^e(\cdot) \stackrel{Thaw}{\longrightarrow} \Unfreeze f^e(\cdot).
\end{equation}

\begin{equation}
\Freeze f^e(\cdot) \stackrel{Thaw}{\longrightarrow} \Unfreeze f^e(\cdot).
\end{equation}

% During the experiments, after the detection head is fine-tuned by Freeze, the entire model is further trained for $500$ iterations for $1$ shot and $2000$ iterations for $2$ and $3$ shots.
 
% \begin{figure*}
%     \centering
%     \includegraphics[width=0.9\textwidth]{pictures/final_hamster.png}
%     \caption{Examples of few-shot video object detection on hamsters (fine-tuned by Freeze). blue bounding boxes are correct detection and red bounding boxes are wrong detection. The first row shows the 1-shot novel-class video used in fine-tuning. The second and third row show the different detection results when the base dataset is weak and strong respectively.}
%     \label{fig:hamster}
% \end{figure*}

\subsection{Results}

\begin{figure}[tbp]
    \centering
    \includegraphics[width=0.45\textwidth]{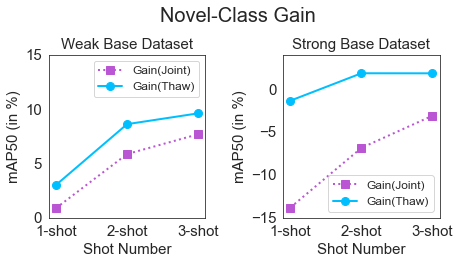}
    %\caption{Novel-class mAP50 gain (in \%) from the component of unfreezing the feature extractor in Joint and Thaw (comparing with Freeze). We can clearly see the gain is increasing with higher shots, and the gain of our Thaw improves over Joint by a large margin. }
      \vspace{-0.1in}
    \caption{Novel-class mAP50 (in \%) improvement of Joint and Thaw over Freeze. The gain is from the component of unfreezing the feature extractor. Clearly,  the gain is increasing with more shots, and  our Thaw improves over Joint by a large margin.}
    \label{fig:novelgain}
          \vspace{-0.2in}

\end{figure}

The experimental results are shown in Table \ref{tab:Thaw}. When the base dataset is \textbf{weak}, it is apparent that our proposed Thaw outperforms Freeze and Joint by a large margin for novel classes. As described in the previous section, Joint outperforms Freeze substantially in this case (see Table \ref{tab:weakandstrong}). Our proposed Thaw is even better, which demonstrates that it can balance the overfitting and insufficiency problems.

 When the base dataset is \textbf{strong}, Thaw performs comparably to Freeze. We should note that Joint performs poorly in this case as also mentioned in Table \ref{tab:weakandstrong}. Our simple method improves Joint by a large margin.

For the base-class performance, our Thaw also significantly outperforms Joint, demonstrating this simple technique significantly alleviates the base-class overfitting problem. Note that the unfreezing part in Thaw can aggravate base-class overfitting problem, so it is expected that Freeze performs the best on the base classes (mentioned in section 5.3).

\noindent{\bf Number of shots vs. insufficiency/overfitting:} When the feature extractor is unfrozen, more shots during few-shot adaptation can simultaneously alleviate novel-class insufficiency and overfitting problems. To better illustrate this phenomenon, we compare the improvement of Joint and Thaw over Freeze on novel classes. The novel-class gain of Joint or Thaw over Freeze is calculated as
\begin{equation}
    {\rm Gain_{Joint/Thaw}}= {\rm mAP50_{Joint/Thaw}}-{\rm mAP50_{Freeze}}.
\end{equation}
The gain can be attributed to the unfreezing feature extractor.  In Figure \ref{fig:novelgain}, it is clear that the novel-class gain increases with the increase of the number of shots.

% \begin{table}[tbp]
%     \centering
%     \caption{Ablation study on two types of classifiers on strong base dataset and novel-base split A.}    
%     \begin{tabular}{c|c|ccc}
%     \hline
%       \multirow{2}*{Method} & \multirow{2}*{Classifier / Shot}  & \multicolumn{3}{c}{Novel-Class mAP50}  \\
%       %\cline{3-11}
%         ~ & ~ & 1 & 2 & 3  \\
%         \hline
%         % \multirow{2}*{Joint} & Fully-connected   & 19.37 & 34.72  & 40.81   \\
%         % ~ & Cosine  & \textbf{21.69}  & \textbf{39.37}  &  \textbf{44.56}  \\
%         % \hline
%         %  \multirow{2}*{Freeze} & Fully-connected    & 39.76  & 47.05  &  51.40 \\
%         %  ~ & Cosine  & \textbf{41.85} & \textbf{50.14} & \textbf{56.00}  \\
%         %  \hline
%          \multirow{2}*{Thaw} & Fully-connected    & 36.91  & 44.62  &  52.86 \\
%          ~ & Cosine  & \textbf{37.81} & \textbf{50.55} & \textbf{56.15}  \\
%         \hline
%     \end{tabular}
%     \label{tab:ablation}
% \end{table}

\begin{table}[tbp]
    \centering
    \caption{Ablation study on the influence of temporal information from the video on the strong base dataset and novel-base split A. Note that {\em 1-shot} video is similar to {\em 15-shot} images. The image-based Freeze (the first row) could be regarded as the implementation of the state-of-the-art few-shot image object detection method \cite{simple} on 15-shot video images.}\vspace{-0.05in} 
    \begin{tabular}{c|c|ccc}
    \hline
       \multirow{2}*{Method} & \multirow{2}*{Format / Shot}  & \multicolumn{3}{c}{Novel-Class mAP50}  \\
       %\cline{3-11}
        ~ & ~ & 1 & 2 & 3  \\
        \hline
        % \multirow{2}*{Joint} & Image-based   & 15.57 &  27.26  &  34.37  \\
        % ~ & Video-based  & \textbf{21.69}  & \textbf{39.37}  &  \textbf{44.56}  \\
        % \hline
        %  \multirow{2}*{Freeze} & Image-based    & 19.73  & 32.24  &  37.56 \\
        %  ~ & Video-based  & \textbf{41.85} & \textbf{50.14} & \textbf{56.00}  \\
        %  \hline
        % Freeze & Image-based   & 19.73 &  32.24  &  37.56  \\
        % \hline
         \multirow{2}*{Freeze} & Image-based    & 19.73  & 32.24  &  37.56 \\
         ~ & Video-based  & \textbf{41.85} & \textbf{50.14} & \textbf{56.00}  \\
        \hline
    \end{tabular}
    \label{tab:ablation2}
    \vspace{-0.1in}
\end{table}

\subsection {Ablation Study and Visualization}

% \noindent{\bf Influence of classifier:}
% We conduct the ablation study of choosing different classifiers in detection head on novel-base split A.  Table \ref{tab:ablation} shows the results of these two classifiers on 1-, 2-, and 3-shot settings with Thaw. We can see that the cosine classifier outperforms the fully-connected classifier in our few-shot video object detection. 

\noindent{\bf Influence of video temporal information:}
Without using the temporal information in the video, the video object detector {\it degenerates} to an image object detector. In this case, given few-shot videos, all the key frames are used separately for fine-tuning the image object detector. Thus 1-shot in our video object detection is equivalent to $15$-shot in image object detection. To study this further, we conduct an ablation study using novel-base split A.  We use Freeze here since image-based Freeze could be considered as the state-of-the-art transfer-learning based few-shot image object detection method \cite{simple}. The results in Table  \ref{tab:ablation2} clearly demonstrate the importance of using video information rather than single image information and the meaning of few-shot learning for video object detection. This could also be seen as a comparison between our few-shot {\bf video} object detection and state-of-the-art few-shot {\bf image} object detection.
%This supports that our proposed problem few-shot video object detection has great contribution in practice.

\noindent{\bf Visualization of few-shot video object detection:}
%We give two examples of 1-shot video object detection results shown in Figure \ref{fig:hamster} and \ref{fig:panda}. The blue and red bounding boxes denote the correct and wrong detection respectively.
An example of 1-shot video object detection results for different methods are shown in Figure \ref{fig:panda}. The blue and red bounding boxes denote the correct and wrong detection, respectively. 
%For the detection results on hamsters in Figure \ref{fig:hamster}, we can clearly see there are fewer red bounding boxes denoted as squirrels when the base dataset is strong. Better features are learnt from a strong base dataset to split from hamsters and similar animals like squirrels, domestic cats.
From the results, we can see that Joint suffers from the overfitting problem as indicated by the red boxes. For Freeze, there is a mixture of correct and wrong detections. For Thaw, there are more correct bounding boxes compared with Joint and Freeze, highlighting its effectiveness.

% \begin{figure}
%     \centering
%     \includegraphics[width=0.4\textwidth]{pictures/overall.png}
%     \caption{Overall-class mAP50 (in \%) comparison of Joint, Freeze and Thaw with different number of shots and different types of base dataset.}
%     \label{fig:overall}
% \end{figure}

% \subsection{Results}

          \vspace{-0.05in}

\section{Conclusion}

We study a new problem of few-shot learning for video object detection. Specifically, we define the problem, construct a new dataset, and utilize a transfer-learning framework for solving this problem. We show that it is not `frustratingly simple' when few-shot learning meets video object detection as insufficiency and overfitting problems are revealed from extensive experiments on our designed weak and strong base datasets by one baseline method (Joint) and one state-of-the-art method (Freeze). Finally, a simple yet effective method called Thaw is naturally developed to validate our analysis and trade off the observed insufficiency and overfitting problems. Our work leads to significantly improved novel-class performance on the weak base dataset and 
competitive novel-class performance on the strong base dataset, while maintaining high base-class performance in few-shot video object detection.
          \vspace{-0.05in}

\section*{Acknowledgement}
This research was funded in part by the New York State Center of Excellence in Data Science. %, an Empire State Development-designated Center of Excellence。 


%\section*{Acknowledgment}




%\bibliographystyle{IEEEtran}


% \end{thebibliography}
\bibliographystyle{IEEEtran}
\bibliography{IEEEabrv}

\end{document}